# Video recordings and play behaviour annotations of group-housed dairy calves on two Dutch farms

Haiyu Yang[1,*], Heidi Lesscher[2], Enhong Liu[1] & Miel Hostens[1]

[1] Department of Animal Science, College of Agriculture and Life Sciences, Cornell University, Ithaca, NY, USA.

[2] Department of Population Health Sciences, Faculty of Veterinary Medicine, Utrecht University, Utrecht, The Netherlands.

* Corresponding author: Haiyu Yang (hy625@cornell.edu; ORCID 0009-0007-5125-4958).

## Abstract

Play is used as an indicator of positive welfare in dairy calves, but openly available video of group-housed calves with individual behaviour annotations is scarce. We describe video from four camera deployments of about one week each in group pens of dairy calves on two farms in the Netherlands: a research and teaching dairy farm (2024 and 2025) and a commercial dairy farm (2024). The recordings comprise 1,173 video files spanning 651 h. For eight observation days, trained observers coded calf behaviour continuously in Noldus The Observer XT with an 18-item ethogram of play and other behaviours, covering 104 h of video, 90 h of which have every calf in the pen coded. The release also contains 1.1 million decoded frames from seven annotated sessions, a per-frame timestamp read by optical character recognition from the camera clock burned into each frame, per-calf segmentation tracks and per-calf frame-level play labels. The data support studies of calf behaviour and welfare and the development and evaluation of automated behaviour recognition.

## 1 Background & Summary

Play in young cattle comprises locomotor elements such as galloping, leaping and bucking, social elements such as frontal pushing and mounting, and play with objects. It is performed mainly when the animal's immediate needs are met and is suppressed by low energy intake, weaning, painful procedures and disease, which is why it is used as an indicator of positive welfare[1–4]. Calves also play more when they have more space[5–7], and space allowance for calves is under discussion in European animal-welfare policy[8]. Because play is brief and makes up only about 1% of a calf's time budget[7], measuring it requires continuous observation of identified individuals over long periods. This is labour-intensive when done by direct observation or from video, and leg-mounted accelerometers, the main automated alternative, capture movement without the visual context needed to distinguish play types[9,10].

Several open video or image datasets of cattle behaviour exist. They include multimodal sensor and camera recordings of lactating dairy cows[11], video clips of five daily behaviours of housed cows[12], short clips of grazing cattle annotated for eleven behaviours[13] and a one-hour recording of eight calves for detection, tracking and identification[14]. A long-term recording of a pig pen is available for studies of behaviour change over weeks[15]. These resources target adult cattle, short clips or non-behavioural tasks, and none of them provides week-long recordings of group-housed calves with continuous, per-calf annotation of play.

The dataset described here was collected within an observational study of play behaviour in group-housed dairy calves. It comprises four camera deployments on two Dutch farms whose owners agreed to open release: Farm A, a research and teaching dairy farm, recorded in April 2024 (pilot), June 2024 and May 2025, and Farm B, a commercial dairy farm, recorded in June 2024 (Table 1, Fig. 1). The release contains (i) all 1,173 playable source video files (651.3 h); (ii) the behaviour event logs of eight annotated observation days, in which trained observers coded the state of every calf in the pen continuously in Noldus The Observer XT, with event times recorded to the video frame, using an 18-item ethogram (89.96 h of video with all calves coded, plus 13.78 h of a pilot day with one focal calf); (iii) 1,106,110 decoded frames from seven sessions that cover six of the annotated days; (iv) a frame index giving, for every decoded frame, the wall-clock time read by optical character recognition (OCR) from the camera clock burned into the image; (v) per-calf bounding boxes and segmentation contours for the calves that were tracked; and (vi) per-calf, per-frame play labels for the calves processed by the labelling pipeline. The camera clock is the time base of the release because the camera recording drops frames, so a frame's position in the file does not give its time reliably.

Parts of these data were used earlier. A preprint by our group[16] analysed play in relation to space allowance across a wider set of farm visits, most of which are not part of this release, and trained a frame-level play classifier on four of the sessions released here together with two sessions from farms that are not included. A second preprint[17] described the processing pipeline and showed frames from the June 2024 recording at Farm A, and a third[18] used calf image crops derived from these recordings, together with crops from farms not included here, as three of several image classes. None of these works released the data or gave a dataset identifier. This Data Descriptor documents the deposited files[19] and the processing that produced them.

Potential reuses include time budgets and bout structure of play in individual calves, synchrony and social play within groups, validation of accelerometer- or video-based play detectors against continuous manual coding, and the development and evaluation of methods for detection, tracking and fine-grained behaviour recognition of small, fast-moving, frequently occluded animals in long videos under farm lighting, including infrared night video. The frame index and the unannotated days make it possible to apply trained models to week-long recordings.

## 2 Methods

### *2.1 Ethics statement*

The study was approved by the Animal Welfare Body of Utrecht University, The Netherlands (approval number 10818-2024-02) and complied with Directive 2010/63/EU. Behaviour was recorded with fixed cameras and was not manipulated. Farmers consented to camera installation and continuous recording, and were told that they might appear on video if they entered the pen. The owners of both farms consented to the public release of the recordings and annotations under a CC0 public-domain dedication. During the visits the calves also wore leg-mounted accelerometers for a separate analysis[16]; the sensor data are not part of this release.

### *2.2 Farms, pens and animals*

The Farm A recordings were made in one group pen of calves during three visits (Table 1). The April 2024 visit (recording FarmA_R1) was a pilot of the recording and sensor procedures; the pen held four calves. The June 2024 visit (FarmA_R2) recorded the same pen with four calves, three of which had also been recorded in April. The May 2025 visit (FarmA_R3) recorded two calves of a later cohort. The Farm B recording (FarmB_R1) was made in one group pen with five calves in June 2024. In the main study, farms were selected for group housing on straw bedding, groups of at least two calves, farmer consent to continuous recording, and pen layouts that allowed full camera coverage[16]. Calves that were clearly ill (health category 4) were not enrolled, and calves vaccinated or dehorned in the preceding two weeks were excluded[16].

At the start of each visit, the observers noted each calf's coat markings, photographed the calves and, where calves looked alike, marked them with coloured livestock spray. In the released files, calves are identified by the names used in the annotation software (a coat description plus the last four digits of the ear tag at Farm A, and pen number, spray mark and coat colour at Farm B); `metadata/calves.csv` maps every spelling to one identifier (12 calves in total).

### *2.3 Video recording*

Each pen was filmed with a Bascom camera system (Bascom Cameras B.V., Nieuwegein, The Netherlands) mounted on the ceiling, a pole or another structure so that the whole pen was visible (Fig. 2)[16]. The cameras recorded without audio and switched to infrared (black-and-white) video in low light. Parts of some views are blacked out by privacy zones set in the camera software. Video was written to SD cards (128 or 256 GB) in the camera units as MP4 files of about 34 min (median 33.7 min), collected at the end of each visit[16] and copied to cloud storage. Every frame carries an on-screen overlay with the camera's date and time to the second and a channel label.

Table 1 lists the camera settings per visit, read from the MP4 headers of every file. The April 2024 camera recorded 2,304 × 1,296 pixels, mostly HEVC with variable frame timing (mean 25.01 frames $s^{-1}$); the June 2024 camera recorded 1,280 × 720 pixels at 25 frames $s^{-1}$, mostly H.264; the May 2025 camera recorded 1,920 × 1,080 pixels at 12 frames $s^{-1}$ (H.264). The April 2024 recording covers about 05:40–24:00 each day, with no recording on 20–21 April; the other three deployments are near-continuous 24-h recordings (Fig. 1). File names encode the recording code, the channel and the start and end times (`FarmA_R2_ch04_20240611_060936_064320.mp4`; April 2024 files use hyphens, `FarmA_R1_ch01-20240424-054842-061639.mp4`). The channel in the file names does not always match the overlay label (Table 1).

The camera recording drops frames (Technical Validation). Consecutive frames in a file are therefore not always the nominal 1/25 s (or 1/12 s) apart, and a frame's wall-clock time cannot be computed from its index or media timestamp. All frame times in this release come from the burned-in camera clock (see *Frame timestamps from the burned-in camera clock*).

### *2.4 Behaviour annotation*

Behaviour was coded from the video in The Observer XT version 16 (Noldus Information Technology, Wageningen, The Netherlands)[20], one Observer XT observation per video file. For every calf in the pen, the observer coded the calf's current state with start and stop events using the ethogram in Table 3; event times are recorded to the video frame (1/25 s, or 1/12 s for the May 2025 camera). The ethogram contains 13 play behaviours, grouped as locomotor (gallop, leap, leap sideways, buck, kick, jump, turn), social (frontal pushing, mount), object (butting fixtures, head-butting a movable object), straw-related (butting or rubbing straw) and play-related (head shake), and five other categories: rest behaviour (any behaviour that is not play), out of view, milk feeding, management (a person in the pen) and individual (a person or other animal visible in front of the pen). Play definitions were adapted from Jensen et al.[5] and the management and out-of-view categories from Gladden et al.[10]. Head shakes were scored as play only directly before or after another play behaviour, and kicks only in a play or social context; kick was a separate behaviour but was scored together with buck when both occurred simultaneously (buck-kick; Table 3).

Observers had practised scoring with the ethogram before coding. Video was watched at normal or half speed whenever a calf was playing and at 2–4× speed when all calves were lying down. In the main study, coding focused on the second day of each visit to minimise effects of the research team's presence[16]; the released sessions also include the first and later days of some visits (Table 2, Fig. 1). Each observation was coded by one observer. Five observers coded the released sessions; they are identified by the pseudonymous codes O1–O5.

Table 2 lists the annotated sessions. On the pilot day at Farm A (18 April 2024) only one focal calf was coded in detail; the other calves carry only placeholder rest states. The seven other annotated days have every calf in the pen coded. The observation start time (`Actual date and start time`) was entered by hand; the video time of an event is this start time plus the event's `Time_Relative_sf`, the event's position in the video file. The export's absolute-time columns hold the observer's computer clock at the time of coding, not the video time.

The authors checked the exported event logs and corrected them before any further processing. They removed six rows of inactive placeholder subjects that carried only rest states, unified the name of one calf that a second observer had spelled differently, corrected three mistyped observation start dates (in `FarmA_2024-04-24`, `FarmA_2024-06-12` and `FarmB_2024-06-19`), merged the partial exports of two Farm B days and removed the suffix ' (Inactive)' that the export had added to every calf name in the last part of 19 June 2024. No behaviour code, event time or duration was changed; `annotations/README.md` lists every change. These corrected exports are the input of all later processing. For release, observer initials in observation names were replaced with the codes O1–O5, the columns of all exports were put in one order, and two columns were added: `calf_id`, the dataset-wide calf identifier, and `observation_start`, the observation start used for event times. `observation_start` is the typed start time, or, where that is missing or mistyped (24 of the 34 observations of `FarmB_2024-06-20` and one pilot observation), the start time in the observation name, which is the name of the video file. The free-text comments of 19 June 2024, which were lost when the partial exports were merged, were restored row by row from those exports.

### *2.5 Frame decoding*

Six of the eight annotated days were decoded to still frames, as seven sessions because 11 June 2024 was decoded in two parts; the April 2024 pilot day and the third Farm B day (20 June 2024) were not decoded (Table 2). For each session, the session's video files were read in order with OpenCV, and every fifth frame of each file (counted from the start of that file) was written as a JPEG at native resolution with OpenCV's default quality (95). Frames are numbered with a seven-digit index that continues across the files of the session (0000001.jpg, 0000002.jpg, …) and are stored in folders of 3,000 frames (video1 = frames 1–3,000, video2 = 3,001–6,000, …). This gives about 5 frames $s^{-1}$ (2.4 frames $s^{-1}$ for FarmA_R3) and 1,106,110 frames in total (Table 2). The mapping from frame index to source file was not stored during decoding.

Three session-specific details affect frame coverage. In FarmA_2024-06-11_AM, frames 0000001–0015000, which covered the recording from 06:09:36 up to the first retained frame, were deleted before this release; the remaining frames keep their original indices, and the overlay of the first retained frame (0015001) reads 06:59:40. The June 2024 file covering 16:46:46–17:20:30 on 11 June was truncated (no MP4 index) and was skipped, so FarmA_2024-06-11_PM has no frames for that interval, which was also not annotated; the file is not part of the release. The decode list for FarmB_2024-06-18 covered 13:03:18–23:06:16, so the annotated intervals 12:51:15–13:00:23 and 23:06:16–23:59:56 (about 63 min) have no frames. The source videos of the deleted and undecoded intervals of FarmA_2024-06-11_AM and FarmB_2024-06-18 are released.

### *2.6 Frame timestamps from the burned-in camera clock*

Because frames are dropped, the time of each decoded frame was read from the camera clock in the image by OCR (Fig. 3). The procedure, implemented as an Apache Spark[21] job, crops a fixed region of interest (ROI) around the clock (Table 6), converts it to grey scale, binarises it with Otsu's threshold and reads it with Tesseract[22] (page segmentation mode 6). Full stops and spaces are removed from the text (the space is kept for the April 2024 camera), and the result is parsed as %d/%m/%Y%H:%M:%S (April 2024: %m/%d/%Y %H:%M:%S, month first). A frame's status is Success if the text parses as a date-time and ParseError otherwise; only Success rows carry a timestamp. Timestamps are local camera time (Central European Summer Time, UTC+2) and are reported at one-second resolution without correction, filtering or interpolation.

For four sessions, the tables produced by the original per-session processing were used unchanged. Three sessions had no usable original table: the original FarmB_2024-06-19 table had been overwritten by the FarmB_2024-06-18 run, and FarmA_2024-06-12 and FarmA_2025-05-12 had never been processed. For these, the same OCR code was re-run with only the paths as parameters, in an environment matching the versions recorded by the original runs (Tesseract 4.1.1 and OpenCV 4.11.0, one OCR thread per process). The FarmA_R3 overlay is at a different position in the 1,920 × 1,080 frames, so its ROI was the only parameter adapted. The frame index (metadata/frame_index/) combines the original and re-run tables and records the source of every row.

### *2.7 Calf tracking*

For each decoded session, calves were located on an early frame with the COCO-pretrained YOLO12x detector[23], run with the default Ultralytics settings (for `FarmA_2024-06-11_PM`, detections were restricted to the class 'cow' with confidence > 0.55), and one box per calf was selected and assigned by hand to the calf's name ('initial prompts'). Each calf was then tracked with SAMURAI[24], a motion-aware tracker built on the Segment Anything Model 2.1 (SAM 2.1, hiera-large video predictor)[25]. Tracking ran separately on each 3,000-frame folder; the first frame of a folder was prompted with the calf's last box in the previous folder. For every calf and frame, the output gives an axis-aligned bounding box and the mask contour. Tracks were not manually corrected, and identity switches were not audited. Tracking was run for 3 of 4 calves on 24 April 2024, 2 of 4 calves on 11 June 2024, all 5 calves on 18 June 2024 (frames 54,001 onwards) and 1 of 5 calves on 19 June 2024 (Table 7). For `FarmA_2024-06-12` and `FarmA_2025-05-12` only the initial prompts exist.

### *2.8 Per-calf frame labels*

Frame-level play labels were produced for the calves and periods processed by the labelling pipeline of the play-behaviour study[16]: three calves on 24 April 2024, two calves on 11 June 2024 and one calf on 19 June 2024 between 06:03 and 11:23 (Table 7). They were made in two steps (Fig. 3).

*Per-calf timeline.* For each calf, the state events of the event log were placed on a 0.1-s time grid (event time = `observation_start` + `Time_Relative_sf`). At each step, the states active for the calf were ranked by start time; the earliest-started state is the step's primary state and the next one its secondary state. Each raw label was mapped to one of three classes: Active Playing (gallop, leap, leap sideways, buck, kick, jump, turn), Non Active Playing (frontal pushing, mount, head shake, head-butting, butting fixtures, butting or rubbing straw, bouncing) and Not Playing (all other labels). The pipeline kept play bouts and their surroundings: Not Playing steps in gaps of more than 1 s between play steps of the calves in the pen were removed, so the timeline does not cover the complete annotated period. The timeline is an intermediate table and is not released; the published code rebuilds it from the event logs (Code Availability).

*Frame labels.* Timeline steps were joined to decoded frames by the frame's OCR second: a step at time $t$ is attached to every frame whose OCR timestamp $s$ satisfies $s \leq t < s + 1$ s. For each calf and OCR second, the primary states of the steps were counted; the most and second most frequent became the calf's `primary_behaviour` and `secondary_behaviour` for that second. The label of the second is Active Playing if either is an Active behaviour, otherwise Non Active Playing if either is a Non-Active behaviour, and otherwise Not Playing.

Every frame of the second receives this label, and the output has exactly one row per frame and calf. Seconds whose primary behaviour is 'out of view' are excluded. Labels therefore exist only for frames with a `Success` OCR reading and for seconds kept in the timeline. For `FarmB_2024-06-19`, whose original OCR table had been overwritten before the labels were built, the frame–second pairs that an earlier processing step had stored from that table were used; they agree with the OCR re-run for all 14,194 labelled frames (Technical Validation).

### *2.9 Personal data*

Farm staff occasionally appear in the footage, mostly as legs or boots in the feed alley; faces are not recognisable. Their presence is coded as 'management' (a person in the pen) or 'individual' (a person or animal in front of the pen). In all released annotation files, observer initials, including those embedded in observation names, were replaced with the codes O1–O5.

### *2.10 Use of AI-assisted tools*

Claude (Anthropic), a large language model, was used to audit the processing code, to write the release and quality-control scripts and the dataset documentation, and to draft this manuscript. The authors checked all code, numbers and text and take responsibility for them.

## 3 Data Records

The dataset is deposited in a public Hugging Face dataset repository[19] (`https://huggingface.co/datasets/Sonam5/Calf-Play-Behavior-Dataset`) under the Creative Commons CC0 1.0 Universal public-domain dedication. The repository revision described here has the DOI https://doi.org/10.57967/hf/10912. Its top-level layout is summarised in Table 4, and the bespoke columns of the main files are defined in Table 5. The farms are identified as Farm A and Farm B, the recordings by the codes `FarmA_R1`, `FarmA_R2`, `FarmA_R3` and `FarmB_R1` (Table 1), and the sessions by identifiers of the form `<farm>_<date>` (for example `FarmA_2024-06-12`), with `_AM`/`_PM` for the two decoded halves of 11 June 2024 (Table 2). All times are local camera time (UTC+2). Timestamp columns created for the release (for example `ocr_timestamp`, `observation_start` and the start and end times in `videos.csv`) use ISO 8601 without a time-zone suffix; the columns of the Observer XT exports keep their original formats.

**Videos** (`videos/<recording>/<YYYY-MM-DD>/<file name>.mp4`; 1,173 files, 1,781.7 GB). Files are grouped by recording (`FarmA_R1`, `FarmA_R2`, `FarmA_R3`, `FarmB_R1`) and by the date on which they start; file names give the recording code, the channel and the start and end times (Methods). FarmA_R3 files that span midnight are named by their end date but filed under their start date. metadata/videos.csv has one row per file with start and end time, file-name span and media duration, size, codec, resolution, frame rate and frame count from the MP4 header, overlap with the annotated windows, the decoded session (if any) and notes on anomalies.

**Decoded frames** (frames/<session_id>/<session_id>_video<NNN>.tar; 372 tar files, 1,106,110 JPEG files, 575.8 GB). Each uncompressed tar holds one original 3,000-frame folder; its members are the JPEG files named by their seven-digit frame index (0015001.jpg), so the shards can be streamed with WebDataset-style readers[26]. Shard video<NNN> holds frames 3,000 × (*NNN* − 1) + 1 to 3,000 × *NNN*.

**Frame index** (metadata/frame_index/<session_id>.csv.gz; 7 files, one row per decoded frame). Columns: session_id, frame_name, chunk (original folder), ocr_timestamp (camera clock read by OCR; empty unless ocr_status is Success), ocr_status and source (original or re-run). A README in the folder documents the OCR method, ROI and source per session. The quality-control tables of the frame index and of the labels are in metadata/qc/ (coverage, plausibility of the Success values, labels on implausible readings, and reproducibility and visual checks of the re-run).

**Annotations** (annotations/; 13 files and a README):

- event_logs/<session_id>.csv (9 files): the behaviour event logs, one per session, as corrected by the authors and pseudonymised (Methods). The 11 June 2024 log is split by observation start into the _AM and _PM sessions. All files have the same columns; the key columns are Observation, Subject (calf name as coded), calf_id, Behavior (raw label), Event_Type (State start, State stop, State point), Time_Relative_sf (seconds from the start of the observation's video file), Duration_sf, Actual date and start time (typed observation start), observation_start and Comment.
- labels/<session_id>.csv.gz (4 files, 57,920 rows): one row per labelled frame and calf, with session_id, frame_name, frame_tar (shard holding the frame), calf_id, subject (calf name as in the event logs), ocr_timestamp (OCR second of the frame), primary_behaviour, secondary_behaviour and label (three-class label).
- README.md: the column definitions of both layers and a table of every correction made to the observers' exports.

**Tracking** (tracking/<session_id>.tar; 7 files, 8.0 GB). Each tar holds <session_id>/masks/video<NNN>/<file per calf>.json (per frame: bounding_box [x, y, width, height] in pixels and the mask contour) and <session_id>/initial_prompts/*.txt (lines calf name: x, y, w, h). For FarmA_2024-06-12 and FarmA_2025-05-12 the tar holds only the initial prompts.

**Metadata** (`metadata/`): `sessions.csv` (one row per session: windows, observations, calves annotated, prompted and tracked, frame counts, index ranges, OCR source and coverage, annotation files and label rows), `videos.csv`, `calves.csv` (one row per calf with every subject spelling), `ethogram.csv` (27 raw label strings with ethogram behaviour, play type, definition, three-class label and event counts per session), `known_issues.md` (every verified anomaly), `qc/` (quality-control tables with a README) and `files_manifest.csv` (path, size and SHA-256 checksum of every file). `LICENSE` holds the CC0 legal code and `README.md` the dataset card.

The repository totals about 2.37 TB, of which the video files account for 1,781.7 GB, the frame shards for 575.8 GB and the tracking archives for 8.0 GB.

## 4 Data Overview

Figure 1 shows, for every recording day, the periods covered by video, by annotation and by released frames. Figure 4 summarises the annotation content. The seven observation days with all calves coded cover 89.96 h of video and 373.5 calf-hours, and the pilot day 13.78 h of one focal calf. Play (active and non-active) occupied between 0.06% and 1.17% of annotated calf-time per session; frontal pushing was the most frequent play behaviour (589 events, 116 min in total), and butting straw (6 events) and turn (8 events) the rarest.

## 5 Technical Validation

### *5.1 File integrity and completeness*

The MP4 header of every video file was read to obtain codec, resolution, frame rate, frame count and media duration (`metadata/videos.csv`). Two files in storage had no MP4 index and could not be played: a 655,360-byte file from 18 April 2024 (11:33–12:15, inside the pilot annotation window, so the pilot observation of 11:33–12:15 has no released video) and a truncated file of 11 June 2024 (16:46–17:20, not annotated). They are not part of the release (`known_issues.md`). The file-name spans of the 1,173 released files total 651.3 h, and their total media duration (648.7 h) is 2.7 h shorter: FarmA_R3 media are 3.6 h shorter, mainly because each FarmA_R3 file spanning midnight contains a recording gap of about 30 min, whereas April 2024 media are 1.7 h longer (see *Frame time base*). A duplicate video and an incomplete upload file were excluded (`known_issues.md`).

Decoded frame counts were checked against the decoding rule without reference to time: the number of frames of a session should equal the sum, over its source files, of $\lceil n/5 \rceil$, where $n$ is the frame count in the MP4 header (`metadata/videos.csv`, columns `n_frames_container` and `decoded_session`). The counts match exactly for `FarmA_2024-04-24`, `FarmA_2024-06-11_PM`,

FarmA_2024-06-12, FarmA_2025-05-12 and FarmB_2024-06-18. FarmB_2024-06-19 has one frame fewer (the missing frame 0095169) and FarmA_2024-06-11_AM 15,001 fewer (the 15,000 deleted frames and the missing frame 0090726). Every decoded frame has exactly one row in the frame index (1,106,110 rows). Before upload, a dry run of the release notebook checked that all source files existed with the expected sizes and built a sample frame shard whose size (1,186,447,360 bytes for 3,000 members) matched the predicted tar size exactly and was byte-identical between two runs. After upload, the SHA-256 checksums that the repository holds for the 1,563 files stored with Git Large File Storage, and the sizes and checksums of the 24 other files, were compared with metadata/files_manifest.csv (1,587 files); all matched.

### *5.2 Frame time base*

The MP4 headers show why frame times cannot be derived from frame positions. For the June 2024 cameras (nominal 25 frames $s^{-1}$), the media duration, which equals the frame count divided by 25, is shorter than the span encoded in the file name by a median of 3.8 s per file of about 34 min for FarmB_R1 (108 of 306 files by more than 5 s) and 3.3 s for FarmA_R2 (95 of 316 files), consistent with dropped frames. The April 2024 camera records with variable frame timing, and its media durations exceed the file-name span in 251 of 263 files. Frame times were therefore read from the camera clock. The same effect applies to the event logs, whose Time_Relative_sf is a position in the video file: for the annotated files the media duration is shorter than the camera-clock span by a median of 1.8–2.9 s per file, depending on the visit (at most 7.7 s), and about 1.8 s longer for the April 2024 camera, so event times computed from the observation start can differ from the camera clock by up to a few seconds towards the end of a file (known_issues.md).

Table 6 reports OCR coverage per session. The proportion of frames with a Success reading ranges from 3.0% (FarmA_2024-06-11_AM) to 99.8% (Farm B). At Farm A in 2024 the overlay is often unreadable in bright daylight: in FarmA_2024-06-11_AM only 12.6% of the minutes of the recording window contain at least one Success reading, and in FarmA_2024-06-11_PM no frame before 17:45 (frames 1–65,049) has one. Annotations in those periods cannot be aligned to frames by the camera clock. The FarmA_R3 and Farm B sessions have a Success reading in every minute.

A Success status means only that the text parsed as a date-time. To quantify misreads, each Success value was compared with the recording window (±5 min) and with the median of its 25 neighbouring Success values on each side (metadata/qc/ocr_plausibility.csv). Between 0% (FarmA_R3) and 3.8% (FarmA_2024-06-11_AM) of Success values fail one of these checks, for example readings with the year 2624 or a date two days off (Table 6). Values are released unchanged; users should screen them (Usage Notes).

The OCR re-run was checked in three ways (`metadata/qc/`). (i) On 40 frames each from three sessions that still have their original tables, the re-run reproduced all 120 status strings and timestamps byte for byte. (ii) On 40 evenly spaced frames per re-run session, the clock was read by eye: 40/40 readings were correct for `FarmB_2024-06-19`; 39 correct and 1 `ParseError` for `FarmA_2025-05-12`; and 15 correct, 1 misread (2 s off) and 24 `ParseError` on legible clocks for `FarmA_2024-06-12`. (iii) For all 14,194 frames of `FarmB_2024-06-19` that carry labels, the frame-index timestamp equals the second used for labelling, which was taken from the original, since overwritten, table.

### *5.3 Annotation quality control*

Every observation was coded by one trained observer, and inter-observer reliability was not measured for these sessions. The released event logs were compared with the observers' original exports on subject, behaviour, event type, relative time and duration: apart from the pseudonymised observation names and the corrections listed in `annotations/README.md`, all values are identical. Event counts per behaviour and session computed by two independent parsers of the event logs (the figure scripts and an earlier profiling script) agreed, and no event failed to parse when the labels were built. Within the annotated windows, 2.8% of calf-time in `FarmA_2024-04-24` and 1.5% in `FarmB_2024-06-19` has no coded state (less than 0.01% in the other sessions); such time should be treated as unlabelled. Known inconsistencies are documented in `known_issues.md` and handled in the metadata tables or in `observation_start`: observation start times missing in 22 and mistyped in 2 of the 34 observations of 20 June 2024 at Farm B (taken from the observation names, which encode the video file), a constant 2-s offset between observation starts and file names on 24 April 2024, spelling variants of calf names (mapped in `calves.csv`) and of labels (mapped in `ethogram.csv`), and two labels that are not defined in the ethogram ('bouncing' and 'play-bouncing', three events at Farm B; mapped to Non Active Playing).

### *5.4 Tracking coverage*

Table 7 lists the tracked calves and 3,000-frame folders per session. Tracks cover every folder of `FarmA_2024-04-24` and `FarmB_2024-06-19`; `FarmA_2024-06-11_PM` lacks folders 1–2 and `FarmB_2024-06-18` starts at folder 19. Because tracks are chained between folders without re-identification, identity switches can propagate; they were not audited, and the tracks should be regarded as automatic pseudo-labels rather than verified annotations.

### *5.5 Frame labels*

The label tables have exactly one row per frame and calf, and the `ocr_timestamp` of every row equals the frame-index timestamp of its frame, which has a `Success` reading (`metadata/qc/labels_ocr_plausibility.csv`). 177 rows, all in `FarmA_2024-04-24`, were excluded as out of view (Table 7). Running the labelling notebook of the published code with the released frame index reproduced all four label tables row for row, and rebuilding the per-calf timeline from the released event logs reproduced the intermediate timelines of `FarmA_2024-04-24` (253,956 rows) and 11 June 2024 (197,313 rows) and, end to end, the 31,776 labels of `FarmA_2024-04-24` without differences. Of the 57,920 label rows, 461 (0.80%) belong to frames whose OCR reading fails the plausibility checks above, 90 of them in `FarmA_2024-06-11_AM` (19.9% of that session's rows); 452 of the 461 are Not Playing. In `FarmA_2024-06-11_AM` these 90 rows are exactly the rows whose OCR second (06:17:11–06:54:50) precedes the first retained frame (06:59:40); the readings are therefore misreads, and the labels of these rows come from the annotation at the misread second. Buck, kick, jump and turn were coded on top of another state and occur in the timeline only as secondary states, so they never determine a label; in 43 label rows (9 in `FarmA_2024-04-24`, 34 in `FarmA_2024-06-11_PM`) the calf has such a secondary Active state in that second but a label other than Active Playing.

## 6 Usage Notes

**Access.** Files can be downloaded individually over HTTPS or with the `huggingface_hub` library; `allow_patterns` selects subsets (for example `annotations/*`, `metadata/*` or one session's `frames/<session_id>/*`), which avoids downloading the 2.37 TB repository. The frame shards can be streamed with WebDataset-compatible readers, including the Hugging Face `datasets` library[27]. The dataset card gives code examples.

**Aligning annotations with frames.** Use `metadata/frame_index/` for frame times; do not compute times from frame indices, file offsets or nominal frame rates, because the camera recording drops frames. Only `Success` rows carry a time. Before use, screen the values against their neighbours, for example by discarding readings more than 60 s from a rolling median of neighbouring readings, as in the plausibility check of the Technical Validation; the release deliberately leaves the values uncorrected. To label frames from the event logs, compute each event's video time as `observation_start` + `Time_Relative_sf` (never the absolute-time columns, which hold the coding computer's clock) and attach it to frames whose OCR second contains it; allow for the drift of up to a few seconds towards the end of a file described in the Technical Validation. Because several frames share one OCR second, labels derived this way have one-second resolution; very short behaviours (for example many kicks and bucks) can fall entirely between decoded frames, particularly in FarmA_R3 (2.4 frames $s^{-1}$).

**Labels.** `annotations/labels/` covers only the calves processed by the labelling pipeline, the seconds kept in the per-calf timeline and frames with a `Success` OCR reading (57,920 calf-frames); most Not Playing time is not labelled. Users who need labels for every calf and every aligned frame should derive them from the event logs as above (the dataset card gives an example) or run the published labelling code on the released tracks. Time inside an annotated window with no coded state, and time outside the annotated windows, is unlabelled, not Not Playing. The three-class mapping of every raw label is in `metadata/ethogram.csv` (`label_3class`); 'out of view' is excluded from the labels. The 'individual' label is a pen-level marker copied to all calves and is not a calf behaviour.

**Training and evaluating models.** Consecutive frames are nearly identical and several frames share one clock second, so frames or calf-frames should not be split at random between training and test sets; split by session, day or time block, with a buffer between training and test data[28,29]. The Farm A sessions of April and June 2024 share calves and the two halves of 11 June 2024 are the same calves on the same day, so they are not independent; treat the three Farm A recordings as one farm in cross-farm evaluation. Play is rare (Table 7), so report balanced accuracy or macro-F1 and per-class recall and precision rather than accuracy.

**Extending the derived files.** The release contains the frames, OCR timestamps, tracks and labels that were produced for the annotated sessions and calves listed in Tables 2 and 7; no further frames were read and no further calves were tracked for this release. Because all 1,173 playable source videos are released, users can decode further frames and run the published OCR, tracking and labelling code (Code Availability) on other periods, sessions or calves. OCR readings of new frames should be checked by eye on a sample, because daylight glare makes the June 2024 Farm A clock unreadable in many frames.

**Scene content.** Calves in neighbouring pens are visible through fences in the June 2024 (behind the far gate) and May 2025 (behind the foreground fence) views; only the pen's own calves are annotated. Vehicles and other animals are occasionally visible outside the pen. The overlay channel label differs from the file names for Farm B (overlay CH3, files ch04), and the June 2024 annotation files and one frame folder are named `Ch01/ch01` although the camera is channel 4 (Table 1).

**Known issues.** `metadata/known_issues.md` lists every verified anomaly, including recording gaps, the two unplayable video files that were removed from the release, deleted and undecoded frame ranges, low OCR coverage at Farm A in 2024, OCR misreads and the label rows that sit on them, annotation start-time problems and the corrections made to the observers' exports, the drift of event times within a file, uncoded calf-time and partial tracking.

**Not included.** The release covers only the two farms whose owners agreed to open release; recordings from the other farms of the observational study are not public. It does not contain the accelerometer data (the sensor-to-calf assignment for April 2024 is unresolved), calf image crops, model weights, the decoded frames of two unannotated days, or intermediate processing tables such as the 0.1-s per-calf timelines, which the published code rebuilds from the released files.

**Relation to the preprints.** The frame labels released here were generated per calf, with one row per frame and calf (Methods); they are not identical to the frame-label tables used for the classifier of the play-behaviour preprint[16]. The calf crops used in the parameter-efficient fine-tuning preprint[18] were cut from these recordings and from recordings on farms not included here; they cannot be reconstructed exactly from this release.

**Updates.** The version reviewed for this Data Descriptor is fixed under its DOI. Any later revision will be released as a new repository revision with a new DOI and a change log in the dataset card.

## 7 Data Availability

The dataset is available from Hugging Face at `https://huggingface.co/datasets/Sonam5/Calf-Play-Behavior-Dataset` (DOI https://doi.org/10.57967/hf/10912)[19] under a CC0 1.0 public-domain dedication, without registration or access restrictions. It comprises the raw videos, decoded frames, OCR frame index, behaviour event logs, per-calf frame labels, calf tracks and metadata tables described in Data Records.

## 8 Code Availability

The code that produced the derived files of the release is available at https://github.com/Sonam525/Play-Behavior-Dataset under the MIT licence (commit dbf5878). It consists of nine notebooks with their parameters in `config/sessions.yaml`: frame decoding (01), OCR of the camera clock (02), the frame index (03), initial prompts with YOLO12x (04), SAMURAI tracking (05), the per-calf 0.1-s timeline (06), the frame labels (07), packaging and upload of the release, including the pseudonymisation of the event logs (08), and the quality-control tables in `metadata/qc/` (09). The notebooks run on Databricks or as plain Python and contain no credentials. The decoding, detection and tracking steps were taken from the video-analysis pipeline of the play-behaviour study[16,17], which is available at

https://github.com/Sonam525/Individual-Behavior-Analysis-with-CV. Software versions: OCR with Tesseract 4.1.1 and opencv-python 4.11.0 (re-runs on Databricks Runtime 15.4 LTS ML with opencv-python 4.11.0.86); labels on Databricks Runtime 17.3 LTS ML with Python 3.12.3 and pandas 2.2.3; tracking with SAMURAI and the SAM 2.1 hiera-large checkpoint; initial prompts with the Ultralytics YOLO12x COCO weights; upload with huggingface_hub 2.0.0 and hf_xet 1.6.0. Versions that were not recorded at run time (the OpenCV version used for decoding, the Ultralytics and PyTorch versions and the SAMURAI commit) are identified as such in the repository README. Parameters are given in Methods and in `config/sessions.yaml`.

## Author Contributions

H.Y.: conceptualization, methodology, software, formal analysis, investigation, data curation, writing – original draft, visualization. H.L.: conceptualization, methodology, investigation, supervision, funding acquisition, writing – review and editing. E.L.: software, writing – review and editing. M.H.: conceptualization, methodology, resources, supervision, project administration, funding acquisition, writing – review and editing.

## Competing Interests

The authors declare no competing interests.

## Acknowledgements

We thank the annotation team of Utrecht University, including Charlotta Kamphuis and Iris de Graaf, for the behavioural coding; the observation protocol and ethogram were developed in C. Kamphuis's MSc thesis research at Utrecht University. We also thank the staff of the two farms for their cooperation.

## Funding

This work was supported by the Cornell Institute for Digital Agriculture and ZuivelNL.

## Figure legends

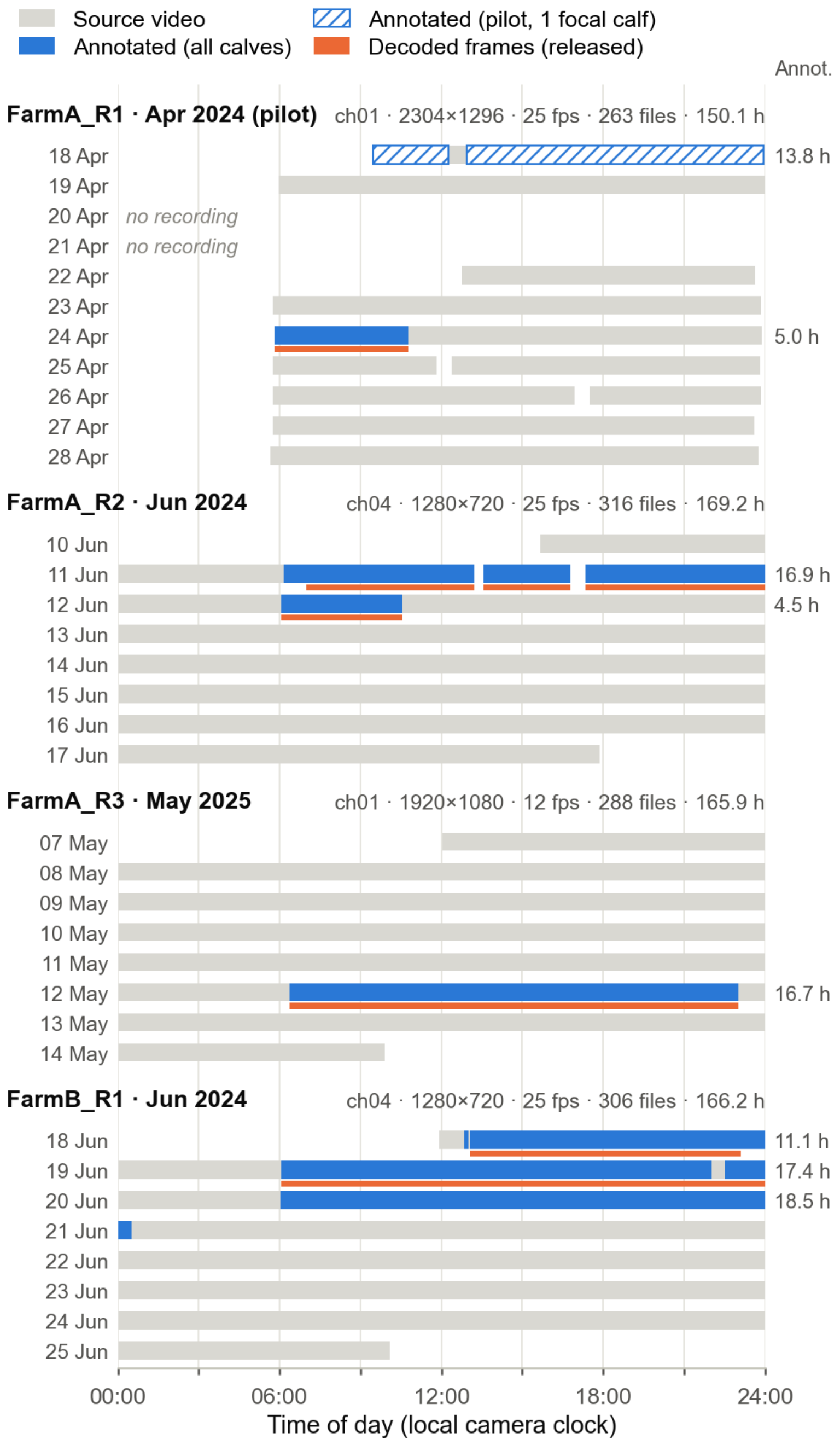

**Figure 1 | Recording calendar of the released video, annotations and decoded frames.** Each row is one calendar day, grouped by recording (Table 1); the x-axis is local camera time. Grey bars, periods covered by video files (file-name start to end); blue bars, periods annotated with every calf in the pen coded; hatched bar, the 18 April 2024 pilot, in which one focal calf was coded; orange strips, periods for which decoded frames are released. Numbers on the right give annotated hours per session, and header lines give recording code and period, camera channel, resolution, source frame rate, number of files and video hours per recording. Gaps in the grey bars are recording gaps (no video on 20–21 April

2024; April 2024 video about 05:40–24:00 only; no video 13:12:54–13:34:27 on 11 June 2024) and two unplayable files that are not released (18 April 2024, 11:33:01–12:15:06; 11 June 2024, 16:46:46–17:20:30). Differences between blue and orange show frames that were deleted (11 June 2024, from 06:09:36 to the first retained frame at 06:59:40), an incomplete decode list (18 June 2024 before 13:03:18 and after 23:06:16) and an annotated day that was not decoded (20 June 2024).

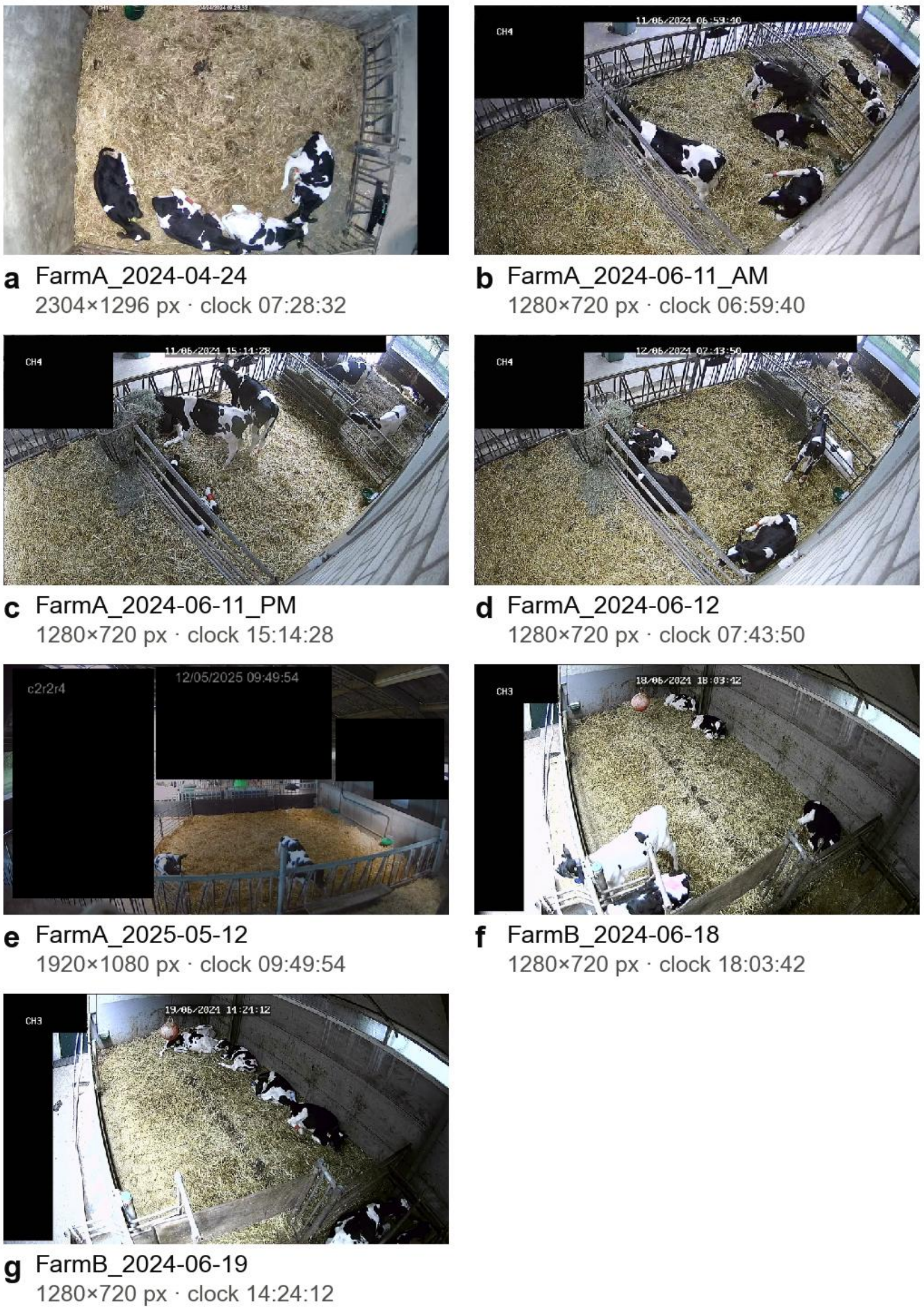

**Figure 2 | Camera views of the seven decoded sessions.** One decoded frame per session, shown unaltered apart from down-scaling to 720 × 405 pixels (Pillow 10.0, Lanczos filter): (a) `FarmA_2024-04-24`, 07:28:32; (b) `FarmA_2024-06-11_AM`, 06:59:40; (c) `FarmA_2024-06-11_PM`, 15:14:28; (d) `FarmA_2024-06-12`, 07:43:50; (e) `FarmA_2025-05-12`, 09:49:54; (f) `FarmB_2024-06-18`, 18:03:42; (g) `FarmB_2024-06-19`, 14:24:12. Labels give session, native resolution and the burned-in clock time. Black areas are privacy zones set in the camera software. Calves in neighbouring pens are visible through the fences in some views (for

example b–d) and are not annotated. Frames were chosen in daylight with calves visible and no person in view.

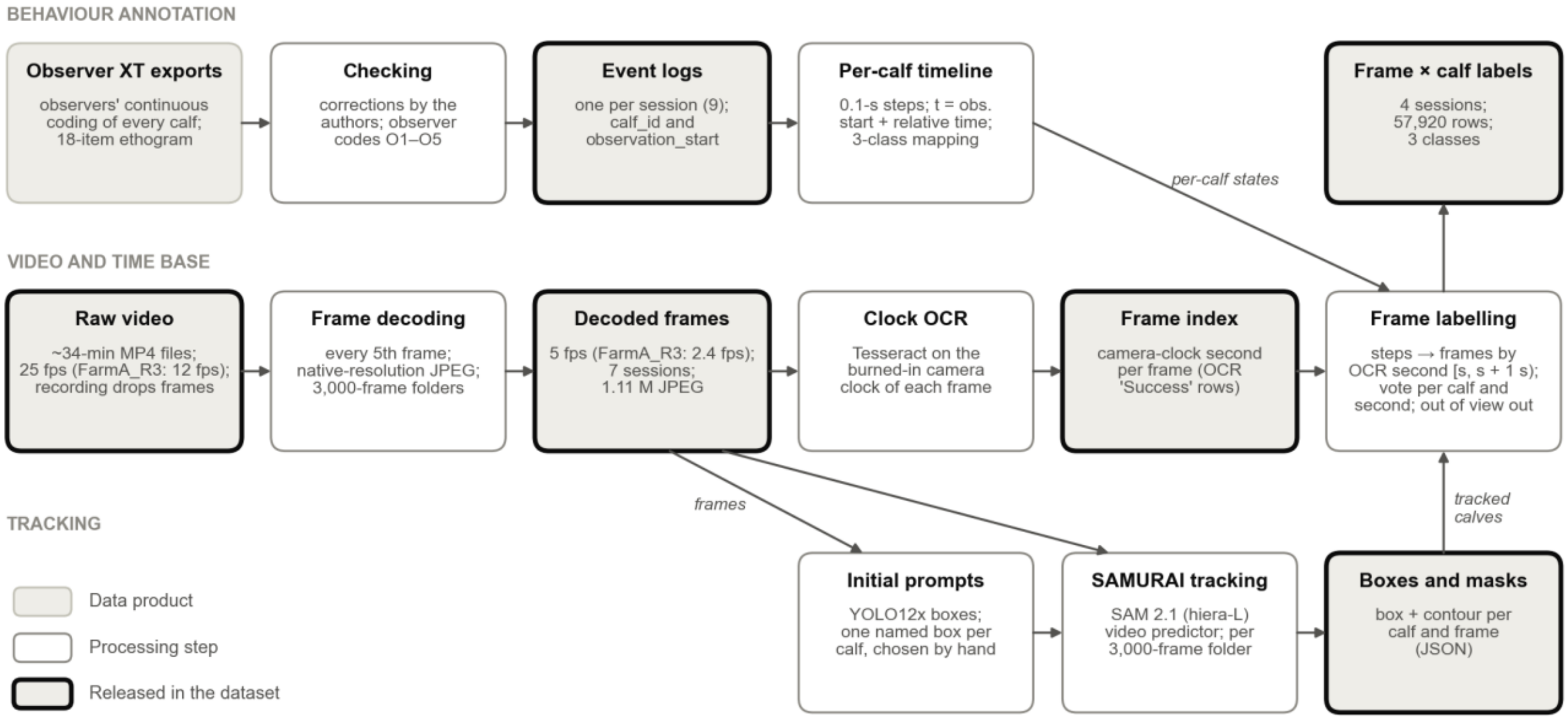

**Figure 3 | Processing workflow.** Behaviour annotation (top): the observers' Observer XT exports were checked and corrected by the authors and are released, pseudonymised, as event logs; for the labelled calves, the state events are expanded to a per-calf timeline at 0.1-s steps and mapped to three classes. Video and time base (middle): every fifth frame of each video file is decoded, and the time of each frame is read by OCR from the burned-in camera clock, because the camera recording drops frames. Tracking (bottom): YOLO12x boxes, assigned by hand to calves, prompt SAMURAI (SAM 2.1) tracking per 3,000-frame folder. In the labelling step, timeline steps are attached to the frames whose OCR second contains them, and labels are voted per calf and second for the tracked calves processed by the labelling pipeline. Shaded boxes are data products and white boxes processing steps; boxes with a bold outline are released in the dataset.

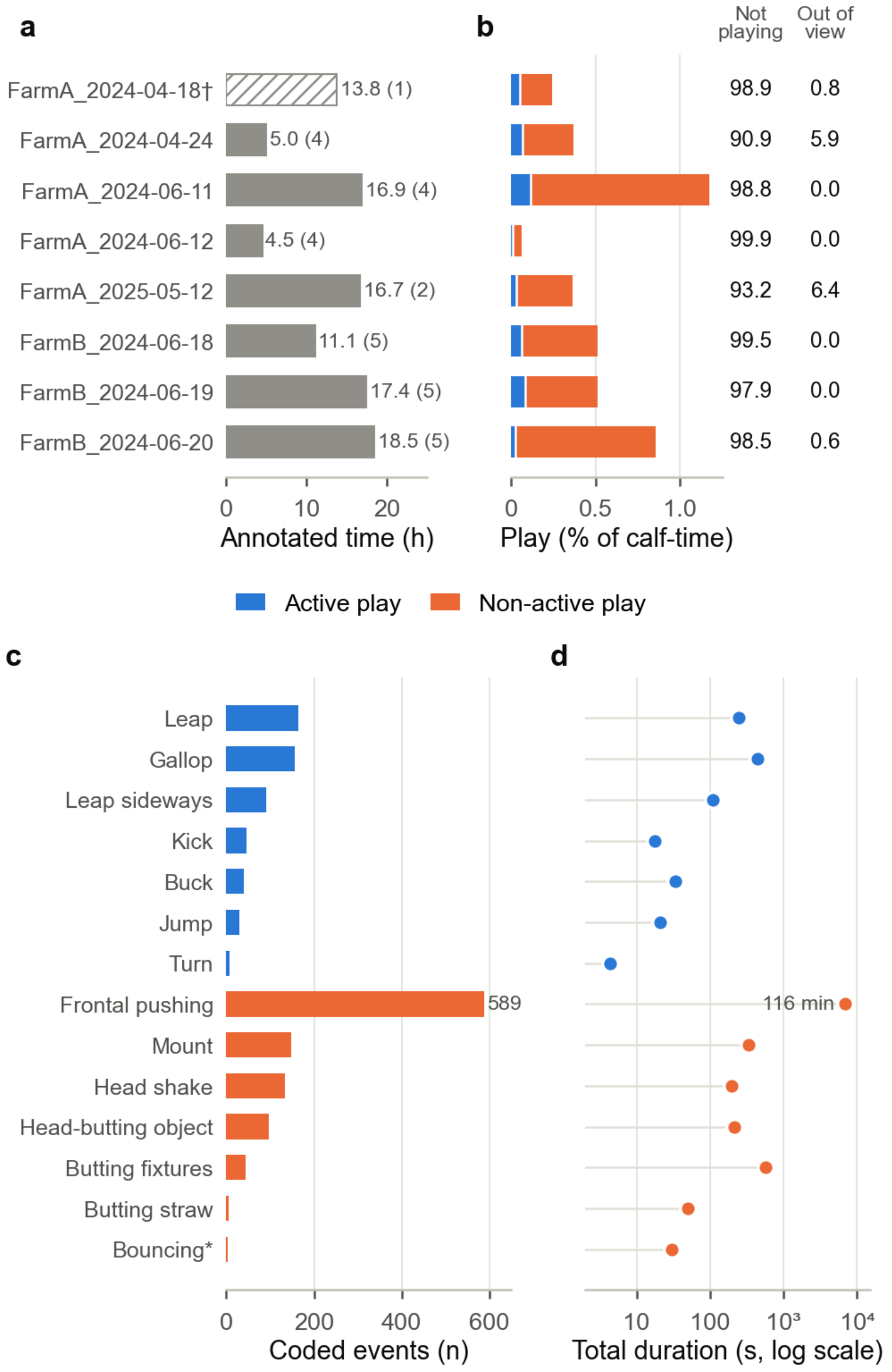

**Figure 4 | Content of the behaviour annotations.** (a) Annotated video time per session (sum of observation lengths; number of calves in parentheses). † The 18 April 2024 pilot, in which one focal calf was coded. (b) Percentage of annotated calf-time in Active play (blue) and Non-active play (orange); the columns on the right give the percentages Not playing and Out of view. Overlapping states were resolved in the priority Active play > Non-active play > Out of view > Not playing. (c) Number of coded events per play behaviour, pooled over sessions. (d) Total coded duration per play behaviour, pooled over sessions (log scale). *'Bouncing' and 'play-bouncing' are labels of three events at Farm B that are not defined in the ethogram. The pen-level 'individual' marker is excluded.

# Tables

**Table 1 | Recording deployments.** Farm A is a research and teaching dairy farm and Farm B a commercial dairy farm. Hours are file-name spans; frame rates, codecs and resolutions are from the MP4 headers of every file.

| Recording | Farm | Dates | View; channel in file names (overlay label) | Codec (files) | Resolution (pixels) | Frame rate ($s^{-1}$) | Files | Hours | Size (GB) | Calves in pen |
|---|---|---|---|---|---|---|---|---|---|---|
| FarmA_R1, April 2024 (pilot) | A | 18–28 Apr 2024; ~05:40–24:00 daily; none 20–21 Apr | top-down; ch01 (CH1) | HEVC (246), MPEG-4 Part 2 (17) | 2,304 × 1,296 | 25 (variable) | 263 | 150.1 | 281.2 | 4 |
| FarmA_R2, June 2024 | A | 10–17 Jun 2024 | oblique; ch04 (CH4) | H.264 (297), MPEG-4 Part 2 (19) | 1,280 × 720 | 25 | 316 | 169.2 | 601.2 | 4 |
| FarmA_R3, May 2025 | A | 7–14 May 2025 | wide-angle oblique; ch01 (c2r2r4) | H.264 (288) | 1,920 × 1,080 | 12 | 288 | 165.9 | 272.8 | 2 |
| FarmB_R1, June 2024 | B | 18–25 Jun 2024 | oblique; ch04 (CH3) | H.264 (306) | 1,280 × 720 | 25 | 306 | 166.2 | 626.6 | 5 |
| Total | | | | | | | 1,173 | 651.3 | 1,781.7 | 12† |

Two unplayable files (18 April 2024, 11:33–12:15; 11 June 2024, 16:46–17:20) are not released and not counted. † Distinct calves; three calves were recorded in April and June 2024.

**Table 2 | Annotated and decoded sessions.** Times are local camera time. Annotated hours are sums of Observer XT observation lengths.

| Session ID | Annotated window(s) | Observations | Annotated (h) | Calves coded | Decoded window(s) | Frames | Frame shards |
|---|---|---|---|---|---|---|---|
| `FarmA_2024-04-18` (pilot) | 09:27:31–12:15:08; 12:57:10–23:55:45 | 20 | 13.78 | 1 focal | not decoded | – | – |
| `FarmA_2024-04-24` | 05:48:40–10:46:52 | 9 | 4.97 | 4 | 05:48:42–10:46:51 | 89,571 | 30 |
| `FarmA_2024-06-11_AM` | 06:09:36–13:12:53 | 13 | 7.05 | 4 | 06:59:40–13:12:54 | 111,860 | 38 |
| `FarmA_2024-06-11_PM` | 13:34:27–16:46:38; 17:20:30–23:59:54 | 19 | 9.84 | 4 | 13:34:27–16:46:46; 17:20:30–23:59:56 | 177,111 | 60 |
| `FarmA_2024-06-12` | 06:03:43–10:33:33 | 8 | 4.49 | 4 | 06:03:43–10:33:35 | 80,884 | 27 |
| `FarmA_2025-05-12` | 06:21:29–23:01:24 | 30 | 16.65 | 2 | 06:21:29–23:01:27 | 143,880 | 48 |
| `FarmB_2024-06-18` | 12:51:15–13:00:23; 13:03:18–23:59:56 | 21 | 11.07 | 5 | 13:03:18–23:06:16 | 180,485 | 61 |
| `FarmB_2024-06-19` | 06:03:43–22:02:00; 22:30:24–23:59:51 | 33 | 17.43 | 5 | 06:03:43–23:59:56 | 322,319 | 108 |
| `FarmB_2024-06-20` | 06:01:35–00:30:26 (next day) | 34 | 18.45 | 5 | not decoded | – | – |

| Session ID | Annotated window(s) | Observations | Annotated (h) | Calves coded | Decoded window(s) | Frames | Frame shards |
|---|---|---|---|---|---|---|---|
| Total (excluding pilot) | | 167 | 89.96 | | | 1,106,110 | 372 |

**Table 3 | Ethogram and three-class mapping.** Definitions are paraphrased from the study ethogram, which was adapted from refs.[5,10].

| Behaviour | Raw label(s) in the logs | Type | Definition | Class |
|---|---|---|---|---|
| Gallop | gallop | locomotor | Runs with a four-beat gait that includes a phase with all four feet off the ground. | Active Playing |
| Leap | leap | locomotor | Lifts both forelegs and jumps forward. | Active Playing |
| Leap sideways | leap sideways | locomotor | Lifts both forelegs and jumps sideways. | Active Playing |
| Buck | Buck | locomotor | Lifts both hind legs so that the weight shifts onto the forelegs. | Active Playing |
| Kick | Kick; kick | locomotor | Kicks one hind leg away from the body to above the tarsus; scored only in a play or social context; scored together with buck when both occur simultaneously (buck-kick). | Active Playing |
| Jump | Jump; jump | locomotor | Jumps vertically without moving forward. | Active Playing |
| Turn | Turn; turn | locomotor | Abruptly changes direction while galloping. | Active Playing |
| Frontal pushing | frontal pushing | social | Pushes its forehead against the head or body of another calf in a playful manner. | Non Active Playing |
| Mount | mount | social | Places both forelegs on the back of another calf. | Non Active Playing |
| Head shake | head-shake | play-related | Shakes or rotates the head; scored only directly before or after another play behaviour. | Non Active Playing |
| Head-butting object | head-butting | object | Pushes its forehead against a movable object such as a ball. | Non Active Playing |
| Butting fixtures | butting fixtures | object | Pushes its forehead against pen fixtures (e.g. feeding equipment, brush), not other calves. | Non Active Playing |
| Butting straw | butting/rubbing straw | straw | Pushes its head into, or rubs its neck in, the straw bedding. | Non Active Playing |
| (not in ethogram) | bouncing; play-bouncing | – | Labels of three events at Farm B; not defined in the ethogram. | Non Active Playing |
| Rest behaviour | rest behavior | other | Any behaviour that is not play (background state). | Not Playing |
| Milk feeding | Milk Feeding | other | Drinks milk; a bout ends when the calf does not return within 60 s. | Not Playing |

| Behaviour | Raw label(s) in the logs | Type | Definition | Class |
|---|---|---|---|---|
| (not in ethogram) | voer krijgen | – | Dutch for 'getting feed'; one event in the 18 April 2024 pilot. | not processed (pilot) |
| Management | Management (someone in the pen); Management (someone in the stalls) | other | A person enters the pen for maintenance or other work. | Not Playing |
| Out of view | out of view | other | The calf's behaviour cannot be seen in the video. | excluded from labels |
| Individual | Individuall | pen-level marker | A person or another animal is visible in front of the pen (point event copied to every calf). | not a calf label |

**Table 4 | Repository layout.** Sizes in decimal GB.

| Path | Content | Format | Files | Size (GB) |
|---|---|---|---|---|
| videos/<recording>/<YYYY-MM-DD>/*.mp4 | source video; file names with the recording code | MP4 | 1,173 | 1,781.7 |
| frames/<session_id>/<session_id>_video<NNN>.tar | decoded JPEG frames, one shard per 3,000-frame folder | uncompressed tar of JPEG | 372 (1,106,110 JPEG) | 575.8 |
| tracking/<session_id>.tar | SAMURAI boxes and contours; initial prompts | tar of JSON and TXT | 7 | 8.0 |
| annotations/event_logs/<session_id>.csv | behaviour event logs, corrected and pseudonymised, one per session | CSV | 9 | < 0.01 |
| annotations/labels/<session_id>.csv.gz | per-calf, per-frame three-class play labels | gzipped CSV | 4 | < 0.01 |
| annotations/README.md | column definitions; corrections made to the observers' exports | Markdown | 1 | < 0.01 |
| metadata/frame_index/ | OCR camera-clock time per frame (<session_id>.csv.gz); README | gzipped CSV, Markdown | 7 + 1 | < 0.01 |
| metadata/*.csv, metadata/known_issues.md | sessions, videos, calves, ethogram, file checksums; known issues | CSV, Markdown | 6 | < 0.01 |
| metadata/qc/ | quality-control tables of the frame index and the labels; README | CSV, Markdown | 5 + 1 | < 0.01 |
| README.md, LICENSE | dataset card; CC0 1.0 legal code | Markdown, text | 2 | < 0.01 |

**Table 5 | Bespoke columns of the frame index, event logs, labels and video table.**

| File | Column | Definition |
|---|---|---|
| frame index | session_id | release session identifier (Table 2) |
| frame index | frame_name | JPEG name, seven-digit frame index within the session (0015001.jpg) |
| frame index | chunk | original 3,000-frame folder (video6); shard |

| File | Column | Definition |
|---|---|---|
| | | `<session_id>_video006.tar` |
| frame index | `ocr_timestamp` | camera clock read by OCR, `YYYY-MM-DDTHH:MM:SS`, local time; empty unless `ocr_status` = `Success`; not corrected |
| frame index | `ocr_status` | `Success` (text parsed as date-time) or `ParseError` |
| frame index | `source` | `original` (original OCR table) or `re-run` |
| event logs | `Time_Relative_sf`, `Duration_sf` | position of the event in the observation's video file, and duration of the state, in seconds |
| event logs | `Actual date and start time` | observation start as typed by the observer (camera clock) |
| event logs | `observation_start` | observation start used for event times (`YYYY-MM-DDTHH:MM:SS.fff`, local time): the typed start, or the start in the observation name where the typed start is missing or mistyped |
| event logs, labels | `calf_id` | dataset-wide calf identifier (`calves.csv`) |
| labels | `subject` | calf name as written in the event logs |
| labels | `ocr_timestamp` | OCR second of the frame used for labelling (equal to the frame index) |
| labels | `primary_behaviour`, `secondary_behaviour` | most and second most frequent primary state of this calf in this second (raw labels) |
| labels | `label` | Active Playing, Non Active Playing or Not Playing |
| labels | `frame_tar` | path of the shard that holds the frame |
| videos.csv | `start_local`, `end_local` | start and end time from the file name, with the true start date |
| videos.csv | `filename_span_s`, `duration_s` | span encoded in the file name; media duration from the MP4 header |
| videos.csv | `codec`, `width`, `height`, `fps`, `n_frames_container` | from the MP4 header |
| videos.csv | `overlaps_annotation`, `annotated_window`, `decoded_session` | whether the file overlaps an annotated window, and which session's; decoded session, if any |
| tracking JSON | `bounding_box`, `contour` | per frame: box [x, y, width, height] and mask polygon, in pixels |

**Table 6 | Camera-clock OCR coverage per decoded session.** ROI in pixels ($x_1$, $y_1$, $x_2$, $y_2$). Implausible: outside the recording window ±5 min or more than 60 s from the median of neighbouring readings.

| Session ID | ROI | Source | Frames | `Success` frames (%) | Minutes with ≥ 1 `Success` (%) | Implausible `Success` values (%) |
|---|---|---|---|---|---|---|
| `FarmA_2024-04-24` | 1010, 0, 1273, 40 | original | 89,571 | 31,081 (34.70) | 100.0 | 851 (2.74) |
| `FarmA_2024-06-11_AM` | 463, 33, 767, 65 | original | 111,860 | 3,352 (3.00) | 12.6 | 128 (3.82) |
| `FarmA_2024-06-11_PM` | 465, 30, 770, 65 | original | 177,111 | 82,682 (46.68) | 49.4 | 12 (0.01) |
| `FarmA_2024-06-12` | 463, 33, 767, 65 | re-run | 80,884 | 28,885 (35.71) | 53.9 | 89 (0.31) |
| `FarmA_2025-05-12` | 725, 20, 1295, 90 | re-run | 143,880 | 143,246 (99.56) | 100.0 | 0 (0.00) |
| `FarmB_2024-06-18` | 465, 30, 770, 65 | original | 180,485 | 180,090 (99.78) | 100.0 | 101 (0.06) |
| `FarmB_2024-06-19` | 465, 30, 770, 65 | re-run | 322,319 | 321,670 (99.80) | 100.0 | 252 (0.08) |

**Table 7 | Tracking and frame-label coverage.** Tracked folders are 3,000-frame folders with SAMURAI output. AP, Active Playing; NAP, Non Active Playing; NP, Not Playing.

| Session ID | Calves coded | Calves tracked (folders) | Calves labelled | Label rows | AP / NAP / NP | Out-of-view rows excluded | Rows with implausible OCR |
|---|---|---|---|---|---|---|---|
| FarmA_2024-04-24 | 4 | 3 (1–30) | 3 | 31,776 | 108 / 314 / 31,354 | 177 | 354 |
| FarmA_2024-06-11_AM | 4 | 2 (6–43) | 2 | 452 | 10 / 34 / 408 | 0 | 90 |
| FarmA_2024-06-11_PM | 4 | 2 (3–60) | 2 | 11,498 | 600 / 5,324 / 5,574 | 0 | 0 |
| FarmA_2024-06-12 | 4 | 0 (prompts only) | – | – | – | – | – |
| FarmA_2025-05-12 | 2 | 0 (prompts only) | – | – | – | – | – |
| FarmB_2024-06-18 | 5 | 5 (19–61) | –† | – | – | – | – |
| FarmB_2024-06-19 | 5 | 1 (1–108) | 1 (06:03–11:23) | 14,194 | 349 / 1,283 / 12,562 | 0 | 17 |
| Total | | | | 57,920 | 1,067 / 6,955 / 49,898 | 177 | 461 |

† Tracked, but not processed by the labelling pipeline, so there are no labels.